\documentclass{llncs}
\usepackage{llncsdoc}

\usepackage{graphicx}
\usepackage{subcaption}
\usepackage[outdir=]{epstopdf}
\usepackage{amssymb}
\usepackage{amsmath}
\newcommand{\mat}{\mathbf}

\usepackage{xcolor}
\begin{document}

\title{Action Learning for 3D Point Cloud Based Organ Segmentation}

\author{Xia~Zhong$^1$, Mario~Amrehn$^1$, Nishant~Ravikumar$^1$, Shuqing~Chen$^1$, Norbert~Strobel$^{2,4}$, Annette~Birkhold$^2$, Markus~Kowarschik$^2$, Rebecca~Fahrig$^2$, Andreas~Maier$^{1,3}$}

\institute{
	$^1$Pattern Recognition Lab, Friedrich-Alexander University Erlangen-N\"urnberg, Germany\\
	$^2$Siemens Healthcare GmbH, Forchheim, Germany\\
	$^3$Erlangen Graduate School in Advanced Optical Technologies (SAOT), Germany\\
	$^4$Fakult\"at f\"ur Elektrotechnik, Hochschule f\"ur angewandte Wissenschaften W\"urzburg-Schweinfurt, Germany\\
}

\maketitle

\begin{abstract}
	We propose a novel point cloud based 3D organ segmentation pipeline utilizing deep Q-learning.
	In order to preserve shape properties, the learning process is guided using a statistical shape model. 
	The trained agent directly predicts piece-wise linear transformations for all vertices in each iteration. 
	This mapping between the ideal transformation for an object outline estimation is learned based on image features.
	To this end, we introduce aperture features that extract gray values by sampling the 3D volume within the cone centered around the associated vertex and its normal vector.
	Our approach is also capable of estimating a hierarchical pyramid of non rigid deformations for multi-resolution meshes.
	In the application phase, we use a marginal approach to gradually estimate affine as well as non-rigid transformations.
	We performed extensive evaluations to highlight the robust performance of our approach on a variety of challenge data as well as clinical data.
	Additionally, our method has a run time ranging from 0.3 to 2.7 seconds to segment each organ.
	In addition, we show that the proposed method can be applied to different organs, X-ray based modalities, and scanning protocols without the need of transfer learning. 
	As we learn actions, even unseen reference meshes can be processed as demonstrated in an example with the Visible Human.
	From this we conclude that our method is robust, and we believe that our method can be successfully applied to many more applications, in particular, in the interventional imaging space.
	
\end{abstract}

\section{Introduction}
	Organ segmentation in 3D volumes facilitates additional applications such as for computer aided diagnosis, treatment planning and dose management.
	Developing such a segmentation approach for clinical use is challenging, as the method must be robust with respect to patient anatomy, scan protocols, and contrast agent (type, amount, injection rate). 
	Additionally, the algorithm should ideally work in real time. 
	3D organ segmentation procedures can be categorized in to model-based, image-based, and hybrid methods.
	For model based methods, where landmark positions are determined by minimizing an energy function, active shape models (\mbox{ASM}), active appearance models (\mbox{AAM}) \cite{wimmer2009generic},
	sparse shape models \cite{zhang2012towards}, and probabilistic atlas models are the most widely used approaches.
	Image-based methods label each voxel in the volume, often referred to as a dense segmentation.
	Graph cut, level set, and deep convolutional neural networks (\mbox{DCNN}), such as the U-net \cite{ronneberger2015u}, fall into this category.
	Hybrid approaches are often the combination of model-based or image-based methods \cite{li2015automatic}.
	While current 3D segmentation systems are dominated by the \mbox{U-net} architecture and its extensions \cite{ronneberger2015u,christ2016automatic}, there are some disadvantages to use a dense segmentation method.
	First, it is difficult to impose geometric constraints utilizing dense labeling, and extra morphological post processing is often required.
	The second disadvantage is the dataset sensitivity, where \mbox{U-net} often requires extensive fine tuning or transfer learning.
	In clinical practice there are often different CT scanners, scan protocols, and contrast agents used for different diagnostic tasks.
	Consequently, a single trained U-net may find it difficult to cope with such data, as input images can have a large variance \cite{chen2017towards}.
	A third disadvantage is the runtime and hardware requirements.
	Although efforts have been made to accelerate the U-net for 3D segmentations \cite{christ2016automatic}, the segmentation of a volume still needs about 100s using a NVIDIA Titan X GPU. 
	With recent advances of reinforcement learning (RL) for medical image processing and network architectures to process point clouds and point sets, we reformulate the segmentation problem in a RL setting.  
	An illustration of the proposed method can be seen in {Fig}.\,{\ref{fig:aperture}}. 
	In the training phase, we augment the ground truth segmentation using affine and non-rigid transformations to generate the state, associated action and reward. 
	Using these data, we trained an agent network to predict the deformation strategy based on the extracted appearance in the current state.
	This trained network is thus used in the testing phase to estimate the deformation of initial estimates iteratively. 
	As our approach is inspired by deep Q-learning, active shape models, PointNet \cite{qi2017pointnet} and U-net \cite{ronneberger2015u}, we refer to it as \textit{Active Point Net}.

\section{Active Point Net}
	In model based methods, the segmentation is formulated as an energy minimization problem.
	Let's denote the volumetric image that corresponds to the underlying ground truth segmentation as $\mat{I} \in \mathbb{R}^3$.
	The objective is to find the system matrix $\mat A = [\mat A_1, \cdots, \mat A_N]$ for the segmentation estimate $\hat{\mat G}(\hat{\mat v}, \hat{\mat e})$ comprising of vertices $\hat{\mat v}$ and edges $\hat{\mat e}$ such that the energy function in Eq.\,\ref{eq:1} is minimized.
		\begin{equation}
			\mat{A} 
			= \arg \min_{\mat{A}} \sum_i \left(\| \mat{I}(\mat{A}\vec{v}_i) - \mat{Q}(\mat{A}\vec{v}_i)\|_2^2 + \lambda\| \mathcal{P}(\mat A_i) \|^2_2 \right)
			\label{eq:1}
		\end{equation}
	where, the regularizer function $\mathcal{P}$ is based on \mbox{ASM} to approximate the transformation $\mat{A}$.
	This optimization problem is computational expensive due to appearance encoding.
	At the same time, the \mbox{ASM} imposes point correspondences onto the segmentation.
	Instead of minimizing the energy function and estimating the shape, we reformulate this problem as a RL problem and learn an agent to predict the transformations $\mat{A}$ directly from a given estimate $\hat{\mat G}(\hat{\mat v}, \hat{\mat e})$ and associated appearance $\mat{\Omega}$.
	In RL, an artificial agent is trained to observe the state of the environment and respond with an action to optimize the immediate and future reward. 
	This process can be modeled as a Q-learning process \cite{watkins1992q} i.e we learn the Q-value of states $\mathcal{S}$, and associated actions $\mathcal{A}$.
	In the context of segmentation, we encode each state $\mat{S}$ as a combination of image features, termed aperture feature $\mat{\Omega}$ and associated vertices $\mat{\hat{v}}$.
	Unlike typical RL problems, during segmentation, the ideal action in each state can be calculated similar to \cite{mnih2015human}.
	Let's denote the ground truth segmentation of the target organ as $\mat{G}(\mat{v}, \mat{e})$.
	The ideal action $\mat{A}^*$ for current states can be calculated by minimizing the energy function
	\begin{equation}
		\mat{A}^* = \arg \min_{\mat{A}} \sum_i \left(\operatorname{d}\left(\mat{A}_i\hat{\vec{v}}_i, \mat G(\mat v, \mat e)\right)^2 + \lambda\| \mathcal{P}(\mat A_i) \|^2_2 \right),
		\label{eq:engergy_opt}
	\end{equation}
	where, function $\operatorname{d}$ denotes the minimal distance between a vertex and the surface of the ground truth.
	Function $\mathcal P$ denotes the penalty term of each affine transformation $\mat{A}_i$, e.g. the smoothness in a local neighborhood.
	Given this ideal action, the reward $r$, associated Q-value given a state and associated action can be formulated as
	\begin{equation}
		r(\mat{S},\tilde{\mat{A}}) = -\|\mat{A}^* - \tilde{\mat{A}}\|_2^2 - \lambda H(\tilde{\mat{A}}\hat{\mat{v}},\mat{v}) 
		\label{eq:loss}
	\end{equation}
	\begin{equation}
		Q(\mat{S},\tilde{\mat{A}}) = r(\mat{S},\tilde{\mat{A}}) + \gamma(\max(Q(\mathcal{\mat{S'},\mat{A'}})))
	\end{equation}
	where $\tilde{\mat{A}}$ denotes the current action constraint based on the search range of current states.
	The function $H$ denotes the Hausdorff distance between the vertices of the deformed estimates and the ground truth.
	Note, that for a given state, there always exists an ideal action such that the term $\gamma\max(Q(\mathcal{\mat{S'},\mat{A'}})) = 0$.
	Therefore, the RL problem is simplified to finding $\tilde{\mat{A}}$ that maximizes the reward $r$ in a given state.
	To process point cloud data, we use an architecture inspired by PointNet \cite{qi2017pointnet} and U-net \cite{ronneberger2015u} to train an agent and use Eq.\,\ref{eq:loss} directly as the \mbox{DCNN's} loss function.
	This reformulation of the segmentation problem has several advantages. 
	First, the computationally expensive energy minimization step can be replaced with a prediction function during the application phase.
	This way, the overall runtime is reduced significantly.
	Secondly, this formulation relaxes the requirement for point correspondences while still keeping the geometric constraints via the smoothness of the system matrix $\tilde{\mat{A}}$.
	This reduces the complexity of the problem.
	Finally, in our reformulation of the segmentation problem, the implementation of data augmentation and utilization of deep learning are straight forward.
	Training data augmentation is performed by arbitrarily deforming the ground truth mesh following its statistical variations as given by the ground truth ASM, calculating the deformation's system matrix w.\,r.\,t.\ an initial segmentation mesh, and subsequently extracting features from the original volumetric image as input for the \mbox{DCNN}.

\subsection{Aperture Appearance Extraction}
\begin{figure}
   		\centering
   		\begin{subfigure}[c]{0.5\textwidth}
   			\includegraphics[width=\textwidth]{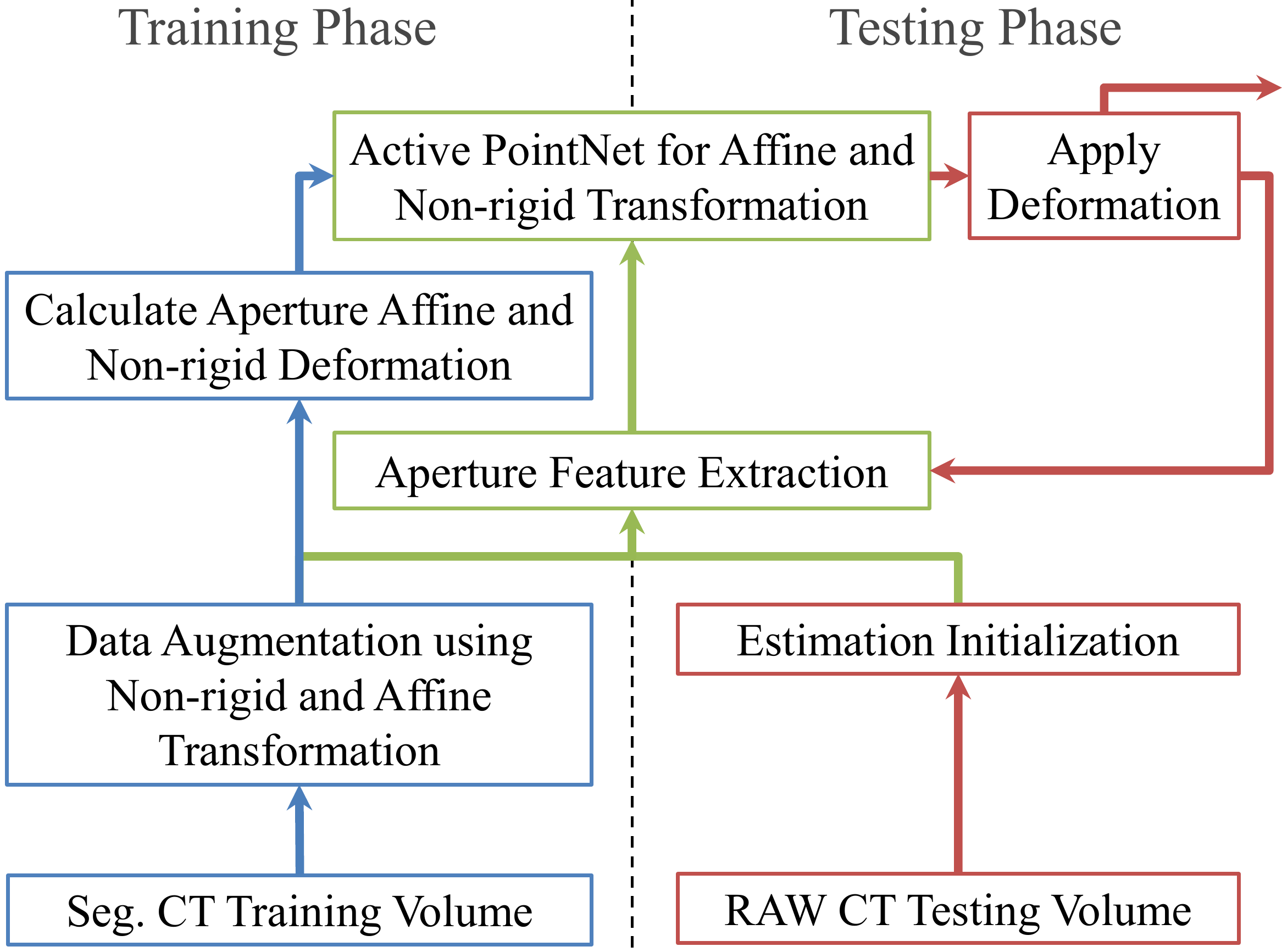}
   		\end{subfigure}
   		\begin{subfigure}[c]{0.42\textwidth}
   			\includegraphics[trim={20 20 60 10},clip,width=\textwidth]{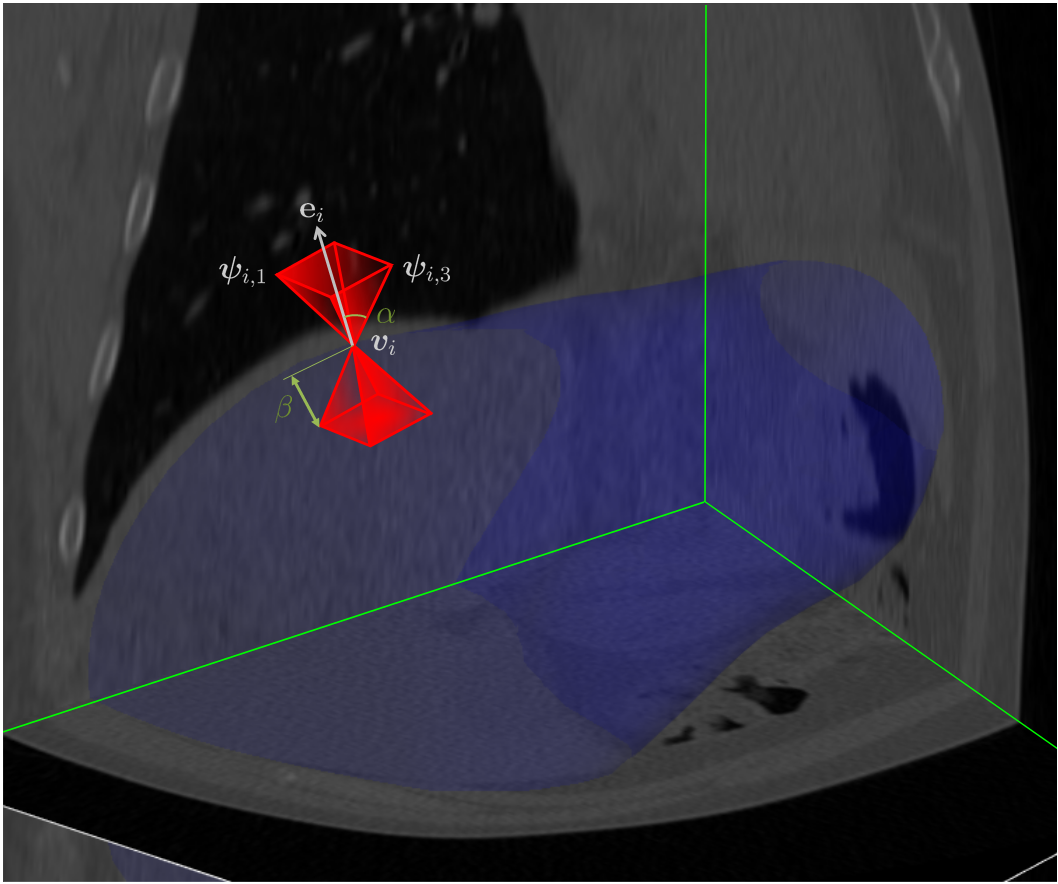}
   		\end{subfigure}

   		\caption{The pipeline of our method (left) and illustration of aperture features (right). The training phase steps are blue, testing phase steps are red and joint steps are green. The aperture boundary is illustrated as a red cone, where the blue surface mesh is the current estimate.}
   		\label{fig:aperture}
   	\end{figure}

	We encode each state, as a concatenation of aperture appearance and associated vertices.
	We extract the surface mesh $\hat{\mat G}$ of an organ and its associated appearance in the 3D volume $\mat I$ by sampling the gray valued reconstructed volumetric X-ray image in a cone of vision.
	We name the cone's extracted features aperture appearance $\mat{\Omega}$.
	For each vertex, a symmetric cone of vision is constructed, which tip is at the vertex location and its rotational axis is defined by the vertex' normal vector. 
	The aperture size $\alpha$ is the cone angle for feature sampling.
	The aperture depth $\beta$ defines the cone's height and maximal capture range of aperture features.
	The aperture density $\vec{\gamma}$ determines the discrete sampling density in depth and size of the aperture features.
	For each vertex $\hat{\vec{v}}_i$, its local coordinate frame \mbox{$\mat{e}_i = [\vec{e}_1, \vec{e}_2, \vec{e}_3]$} is calculated, where $\vec{e}_1$ is the vertex' normal vector. % [Difference between max{e} and vec{e} not visible in PDF! Maybe use different characters]
	The boundary of the cone of vision is constructed with angle $\alpha$ as well as maximal length $\beta$.
	Therefore, we can define the boundary of the aperture for each vertex as \mbox{$\mat{\Psi}(\alpha, \beta, \gamma, \hat{\mat{G}}) = [\mat{\Psi}_i, \cdots, \mat{\Psi}_\text{N}]$}, where \mbox{$\mat{\Psi}_i = [\vec{\psi}_1, \cdots, \vec{\psi}_4]$} denotes the vector of the boundary of the aperture cone. 
	We extract the gray values of associated sampling points in the aperture $\mat{\Theta}$ and concatenate these features with the coordinates of all vertices $\hat{\mat v}$, their normal vectors, and the aperture boundary $\mat{\Psi}$.
	The concatenated features are the aperture appearance \mbox{$\mat{\Omega}(\hat{\mat v}, \mat{\Theta}, \mat{\Psi}\,|\,\mat{I},\hat{\mat G})$}.
	An illustration of aperture appearance extraction is demonstrated in {Fig.}\,{\ref{fig:aperture}}.

\subsection{Aperture Deformation Pyramid}
	For each aperture appearance $\mat{\Omega}$, the action deformation matrix $\tilde{\mat{A}}$ is predicted by the network.
	For each vertex, the associated $\tilde{\mat{A}}$ is restricted by the range of the aperture appearance.  
	In the current implementation, we encode each vertex' deformation as a translation vector.  
	Note that in the special case of an aperture size $\alpha = 0$, %the aperture cone degenerates to the normal vector and 
	the associated deformation model is simplified to a movement along the normal vector.
	Therefore, the non-rigid transformation \mbox{$\tilde{\mat{A}} \in \mathbb{R}^{N\times1}$} describes the signed magnitude indicating the movement along the vertices' normal vectors.  
	We can further expand the single resolution deformation to a multi-resolution model.
	To this end, we use subdivision to increase the resolution of the estimates $\hat{\mat{G}}$ and calculate the associated high resolution system deformation matrix.
	We can iterate this process and form a deformation pyramid for different resolutions.
   	To reduce computational complexity and increase efficiency, we train a single agent for different resolution levels by sub-sampling the mesh and associated deformation in a higher resolution and divide them into batches.
   	The number of vertices in each batch matches the original resolution.
   	We further decompose the deformation into a global affine  and a local non-rigid transformation and use separate agents to predict the global and local action.  
   	
\subsection{Data Augmentation}
	Data augmentation of training data is performed by randomly translating vertices of an organ's surface mesh.
	The appearance of a deformed mesh is extracted using the coordinates of augmented vertices.
	In parallel, the optimal alignment and the deformation model is calculated based on coordinates of the known ground truth mesh.
	To deform the vertices, we train an active shape model of the surface mesh and use its deformation vector for the non-rigid augmentation. % [Could be clearer: Not the vertices, but the mesh is deformed (vertices are translated), what do we call this "initial/current segmentation" mesh to not confuse it with the GT mesh? Do you want to say "alignment model"? or "alignment and deformation model"?]
	To cover a large variety of scenarios in the evaluation of the proposed method, we apply the \mbox{ASM} deformation using all modes to the ground truth surface mesh and subsequently apply a random affine transformation matrix.   
	We also introduce random jittering of each vertices at the end of the augmentation as regularization  to avoid overfitting during the learning process.
	
\subsection{Artificial Neural Network Architecture}
\begin{figure}
		\centering
		\includegraphics[width=0.9\textwidth]{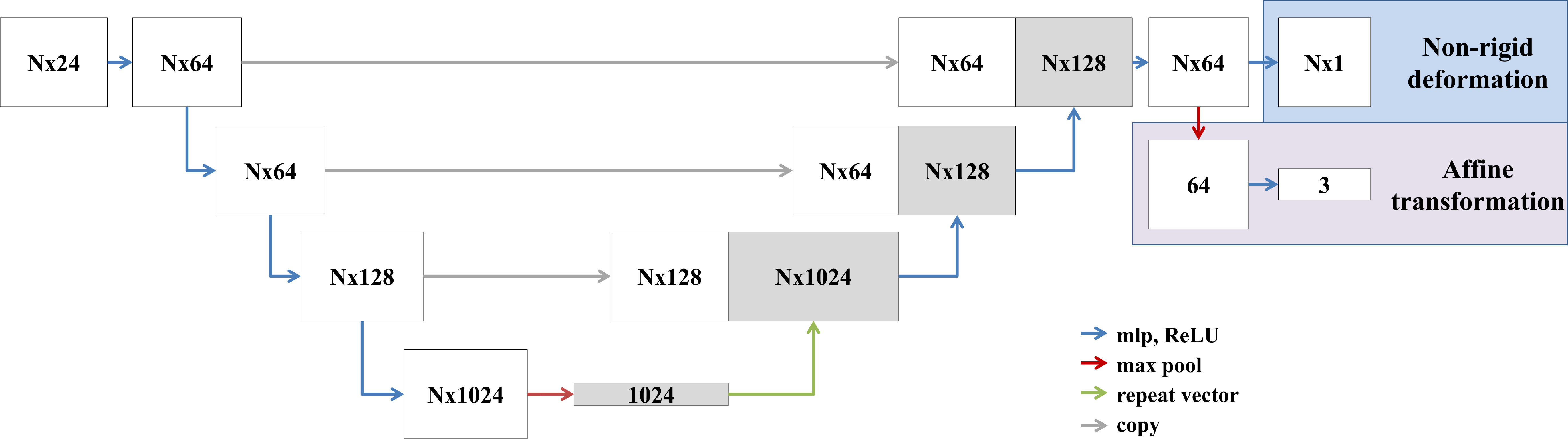}
		\caption{Illustration of the network architecture used to update a current object surface mesh estimate. Featuremaps are represented as boxes.}
		\label{fig:network}
	\end{figure}

	Our proposed network architecture for the global and local action agent is visualized in {Fig}.\,{\ref{fig:network}}.
	This network is inspired by \mbox{PointNet} \cite{qi2017pointnet} and \mbox{U-net} \cite{ronneberger2015u}, where we use a shared multi-layer perceptron (\mbox{MLP}) for intra vertex local feature calculation.
	This way, no assumptions for neighboring vertices are made and therefore, local features are insensitive to the permutation of vertices.
	Subsequently, a pooling function is used to calculate the global feature for all vertices.
	Endorsing the \mbox{U-net} architecture, we aggregate multiple levels of local features along the global features instead of just one type of feature, as in the original \mbox{PointNet}, to estimate the affine and non-rigid transformations.
	Furthermore, we decompose the system matrix into global and local transformations.
	We set the local transformations as the output of the network while we use another pooling layer to estimate the global transformation.
	The global transformation is split into translation, rotation and scaling for balancing the loss function.
	In the application phase, we use a marginal approach by estimating translation, affine transformation, and non-rigid deformation sequentially to further increase the efficiency of our approach.

\section{Evaluation}
	\subsubsection*{Dataset}
	For evaluation, we trained our network using $66$ full-body \mbox{CT} volumes, among which six were used for validation and made $200$ augmentations for each volume.
	We tested our trained networks on Visceral data \cite{del2014visceral}, \mbox{MICCAI} \mbox{LiTS} 2017 challenge data, % https://competitions.codalab.org/competitions/17094
	as well as clinical datasets. % [which challenges, name theHausdorffHausdorffm. Maybe even reference them]
	{Tab}.\,{\ref{tab:segmentation_results}} summarizes number of cases, regions of interest (\mbox{ROI}), modalities, and results.
	The S{\o}rensen-Dice coefficient and the Hausdorff distance were chosen as figure of merit for comparison.

	\subsubsection*{Accuracy and Runtime}
	Our method's results depend on the definition of an initial mesh for the segmentation process.
	Therefore, we randomly initialize the surface mesh ten times for each dataset and evaluate the mean and standard deviation to show the accuracy and robustness of our method. 
	The random initialization comprises a maximum translation of eight centimeters in each direction and a deformation comprising the first five \mbox{ASM} modes. 
	To compare different configurations of our algorithm, we evaluate the liver segmentation using various settings. 
	{Tab}.\,{\ref{tab:segmentation_comparision}} states the segmentation accuracy using degenerated or normal aperture combinations with single resolution or multi-resolution approaches. 
	We also performed a comprehensive evaluation on different organs and different datasets. 
	These results are summarized in {Tab}.\,{\ref{tab:segmentation_results}}.  
	The training and testing was done on a NVIDIA K2100M GPU.

\section{Discussion and Outlook}
\begin{figure}
		\centering
		\begin{subfigure}[c]{0.30\textwidth} % (499 / 567) / (491 / 375) == 0.67215164
			\includegraphics[height=0.18\textheight]{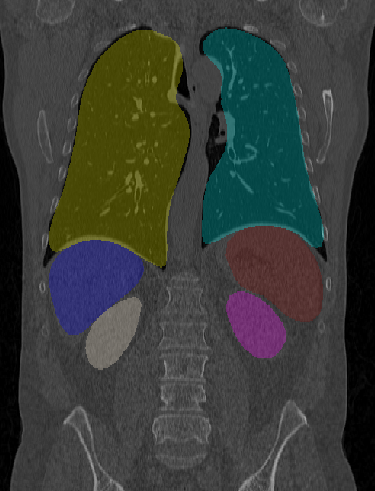} % 375x491
		\end{subfigure}
		\begin{subfigure}[c]{0.33\textwidth}
			\includegraphics[height=0.18\textheight]{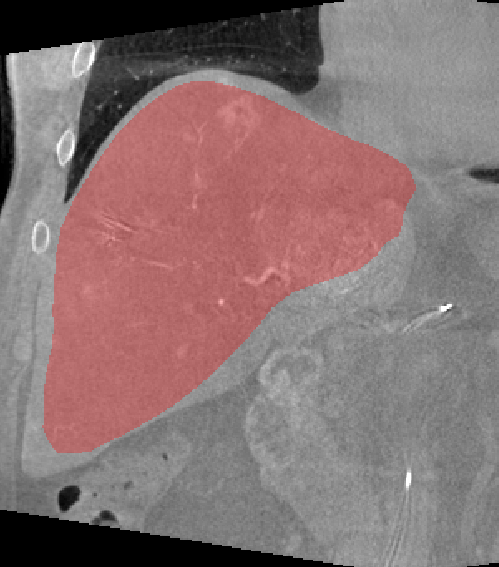} % 499x567
		\end{subfigure}
		\begin{subfigure}[c]{0.32\textwidth}
			\includegraphics[height=0.18\textheight]{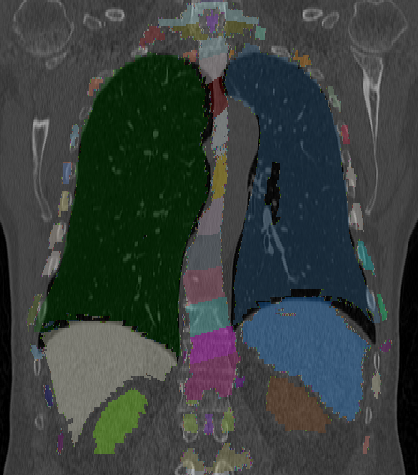} % SAME IMAGE! 499x567
		\end{subfigure}
		
		\caption{The result of multi-organ segmentation of a CT dataset (left), an example of the liver segmentation of a flat-panel CT scan of a liver (middle) and the warped visible human phantom overlay with a CT (right).}
		\label{fig:result}
	\end{figure}

	We have shown that our method is accurate as well as robust for multi-organ segmentation.
	The proposed method is robust w.\,r.\,t.\ initialization and different datasets.
	We also observed that the proposed method is robust w.\,r.\,t.\ CT scan technology as well as scan protocols. 
	We tested our trained network using intra-operative flat panel \mbox{CT} datasets without transfer learning or fine-tuning. 
	Results using our network are shown in {Fig}.\,{\ref{fig:result}}. 
	% An result using our network can be shown in Fig. X. [This should be in Results/Evaluation, not Discussion]
	The proposed method presents a network for organ segmentation utilizing data with different scanning protocols and secondary modalities.  
	Although, the proposed method alone does not outperform state-of-the-art methods trained specifically on the respective data sets, our results come close. 
	The runtime, however, is $20$\,--\,$200$ times faster than comparable methods \cite{christ2016automatic,li2015automatic}, making it an ideal candidate for interventional use where time is of utmost importance.
	\begin{table}
		\begin{center}	
			\begin{tabular}{l c c c}
				\hline
				\hline
				Algorithm & Dice & Hausdorff & time per volume \\
				& [\%] & [mm] & [s] \\
				SR \ $\alpha$ = 0 & 87.8 $\pm$ 3.6 & 30.5 $\pm$ 12.2 & \textbf{0.9} \\
				SR \ $\alpha$ = $\pi$/8 & \textbf{88.6 $\pm$ 4.5} & {27.5 $\pm$ 12.0} & 6 \\
				MR \ $\alpha$ = 0 & 88.4 $\pm$ 3.6 & 29.7 $\pm$ 12.2 & 4.5 \\
				MR \ $\alpha$ = $\pi$/8 & {88.5 $\pm$ 7.3} & \textbf{26.7 $\pm$ 12.4} & 26 \\
				\hline
				\hline
			\end{tabular}
		\end{center}
		\caption{Liver segmentation accuracy and runtime comparison using the normal aperture or degenerated aperture feature combining with single resolution (SR) or multi-resolution (MR)}
		\label{tab:segmentation_comparision}
	\end{table}
	As all point correspondences are modeled only implicitly, we can use our Active Point Net even on unseen data such as the Visible Human (VisHum) \cite{spitzer1996visible}. 
	An example of this application is shown in {Fig}.\,{\ref{fig:result}}.
	Using the VisHum's anatomical model as the initialization, we can easily deform its major organs according to our state-action-function to match the 3D patient data. There is no correspondence between patient, learned update steps, and the VisHum required.
	In addition the method can also be applied to estimate deformation fields from the organ surface deformations using interpolation techniques such as thin plate splines. As such, we believe that the method is extremely versatile and might be useful for numerous clinical applications in the future.
	
	\subsubsection{Acknowledgements}
	We gratefully acknowledge the support of Siemens Healthineers, Forchheim, Germany. 
	Note that the concepts and information presented in this paper are based on research, and they are not commercially available.
	
	\begin{table}
		\begin{center}	
			\begin{tabular}{ l  c  l  l  l  c  c  c  c}
				\hline
				\hline
				Dataset	& Num & \mbox{ROI} & Modality & Organ & Dice & Hausdorff & time per volume\\
				
				& & & & & [\%] & [mm] & [s]\\
				\hline
				LiTS-1* 		& 27 & Liver 		& CT 	& Liver 	& 89.3 $\pm$ 2.4 & 28.4 $\pm$ 8.9 & 0.9 \\
				LiTS-2* 		& 82 & Liver 		& CT 	& Liver 	& 87.9 $\pm$ 4.2 & 35.8 $\pm$ 15.2 & 0.9 \\
				\hline
				Anatomy3 	& 10 & CAP 			& CT 	& Liver 	& 89.5 $\pm$ 1.9 & 26.8 $\pm$ 7.4 & 0.9 \\
				&	 &				&		& Spleen 	& 83.8 $\pm$ 5.3 & 19.4 $\pm$ 11.8 & 0.6 \\
				&	 &				&		& Kidney-L 	& 75.7 $\pm$ 13.7 & 21.4 $\pm$ 11.8 & 0.3 \\
				&	 &				&		& Kidney-R 	& 74.8 $\pm$ 11.3 & 20.2 $\pm$ 8.2 & 0.3 \\
				&	 &				&		& Lung-L 	& 94.2 $\pm$ 1.4 & 17.1 $\pm$ 7.9 & 2.6 \\
				&	 &				&		& Lung-R 	& 94.0 $\pm$ 1.0 & 18.8 $\pm$ 6.8 & 2.6 \\
				\hline
				SilverCopus* & 19 & ThAb 		& CECT 	& Liver		& 88.0 $\pm$ 3.1 & 36.5 $\pm$ 17.5 & 0.9 \\
				&	 &				&		& Spleen 	& 76.3 $\pm$ 13.1 & 35.1 $\pm$ 34.1 & 0.8 \\
				&	 &				&		& Kidney-L 	& 79.8 $\pm$ 11.4 & 22.8 $\pm$ 7.1 & 0.3 \\
				&	 &				&		& Kidney-R 	& 63.6 $\pm$ 19.1 & 32.8 $\pm$ 15.7 & 0.3 \\
				&	 &				&		& Lung-L 	& 94.2 $\pm$ 1.2 & 19.9 $\pm$ 8.5 & 2.7 \\
				&	 &				&		& Lung-R 	& 94.3 $\pm$ 0.8 & 19.3 $\pm$ 5.7 & 2.7 \\
				\hline 
				DECT* 		& 40 & ThAb 		& DECT 	& Liver		& 88.5 $\pm$ 3.6 & 29.7 $\pm$ 12.2 & 0.9 \\
				&	 &				&		& Spleen 	& 83.7 $\pm$ 5.7 & 17.8 $\pm$ 7.9 & 0.7 \\
				%\hline
				%FPCT 		& 10 & Liver 		& FPCT 	& Liver		& \\ 
				\hline
				\hline
			\end{tabular}
		\end{center}
		\caption{Segmentation result and runtime of different organs in different datasets  (chest-abdomen-pelvis (CAP), thorax-abdomen (ThAb)) and modality(CT, contrast enhanced CT (CECT) and dual energy CT (DECT)). The notation * means the segmentation is performed without transfer learning or fine tunning.}
		\label{tab:segmentation_results}
	\end{table}

\bibliographystyle{splncs}

\bibliography{ref}

\end{document}